\documentclass[a4paper,11pt, DIV=12]{scrartcl}

\usepackage[utf8]{inputenc}

\usepackage{csquotes}
\usepackage{epsfig}
\usepackage{amsmath}
\usepackage{amssymb}
\usepackage[binary-units]{siunitx}
\usepackage{amsfonts}
\usepackage{booktabs}
\usepackage{multirow}
\usepackage[toc,page]{appendix}
\usepackage{caption}
\usepackage{subcaption}
\usepackage{algorithm2e}
\usepackage{textcomp}
\usepackage{xspace}
\usepackage{multimedia}
\usepackage{graphicx}
\usepackage{float}
\usepackage{pgfplots}
\pgfplotsset{compat=1.17}
\usetikzlibrary{spy}
\usepackage{authblk}
\usepackage[toc,page]{appendix}
\usepackage[english]{babel}
\usepackage[round]{natbib}
\newcommand{\rs}{Intel® RealSense™ D435\xspace}

\newcommand{\jetson}{NVIDIA® Jetson Nano™ Developer Kit\xspace}

\newcommand{\tierpark}{Lindenthal Zoo\xspace}

\newcommand*\annotatedFigureBoxCustom[8]{\draw[#5,thick] (#1) rectangle (#2);\node at (#4) [fill=#6,thick,shape=circle,draw=#7,inner sep=2pt,text=#8] {\textbf{#3}};}
\newcommand*\annotatedFigureBox[4]{\annotatedFigureBoxCustom{#1}{#2}{#3}{#4}{white}{white}{black}{black}}

\newenvironment {annotatedFigure}[1]{\centering\begin{tikzpicture}
\node[anchor=south west,inner sep=0] (image) at (0,0) { #1};\begin{scope}[x={(image.south east)},y={(image.north west)}]}{\end{scope}\end{tikzpicture}}

\allowdisplaybreaks

\title{Exploiting Depth Information for Wildlife Monitoring}

\author{Timm Haucke}
\author{Volker Steinhage \footnote{\{haucke,steinhage\}@cs.uni-bonn.de}}
\affil{University of Bonn, Institute of Computer Science IV, Friedrich-Hirzebruch-Allee 8, Bonn 53115, Germany}

\begin{document}

\maketitle

\begin{abstract}
    \noindent
    Camera traps are a proven tool in biology and specifically biodiversity research. However, camera traps including depth estimation are not widely deployed, despite providing valuable context about the scene and facilitating the automation of previously laborious manual ecological methods. 
    
    In this study, we propose an automated camera trap-based approach to detect and identify animals using depth estimation. To detect and identify individual animals, we propose a novel method \textit{D-Mask R-CNN} for the so-called instance segmentation which is a deep learning-based technique to detect and delineate each distinct object of interest appearing in an image or a video clip. An experimental evaluation shows the benefit of the additional depth estimation in terms of improved average precision scores of the animal detection compared to the standard approach that relies just on the image information. This novel approach was also evaluated in terms of a proof-of-concept in a zoo scenario using an RGB-D camera trap.
    
\end{abstract}

\clearpage

\section{Introduction}

\begin{figure}[H]
    \begin{subfigure}[b]{\textwidth}
        \centering
        \includegraphics[width=0.625\textwidth]{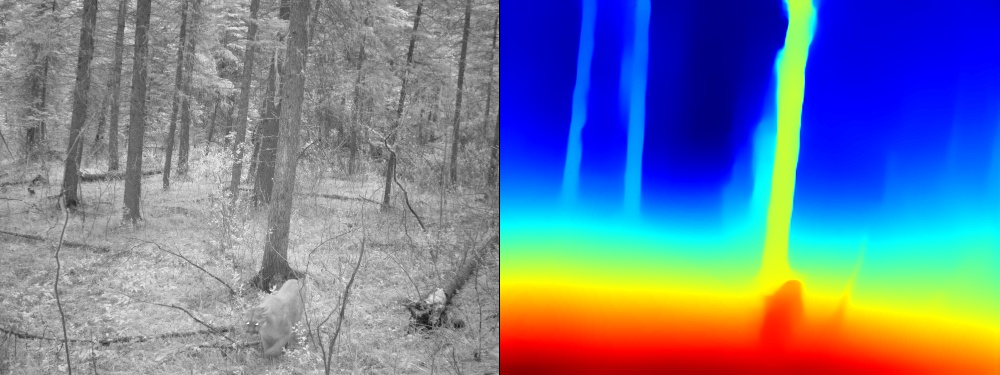}
    \end{subfigure}
    \hfill
%
    \begin{subfigure}[b]{\textwidth}
        \centering
        \includegraphics[width=0.625\textwidth]{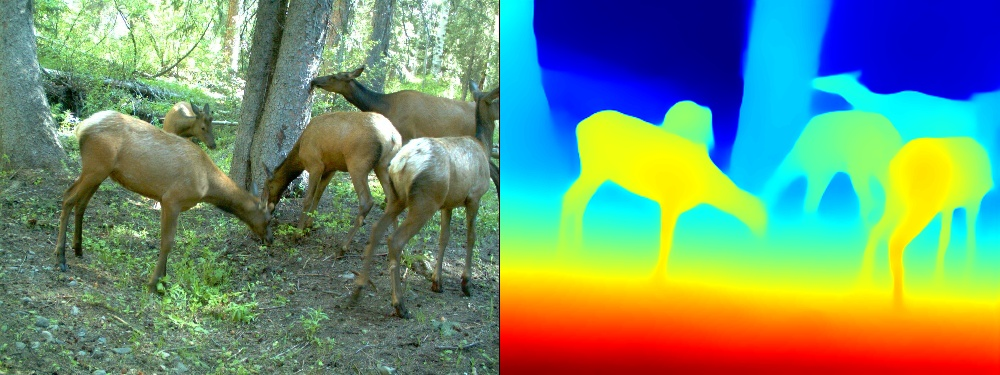}
    \end{subfigure}
    \hfill
    \caption{Depth information supports a more reliable detection of animals at all as well the distinction of individual animals in hordes. Depth information is encoded using heatmaps where distance is highest in blue and lowest in red. Original images are from the Caltech Camera Traps dataset \citep{cct} and depth is estimated using MiDaS \citep{midas}.}
    \label{fig:foreground_background_separation}
\end{figure}

The biodiversity crisis, i.e. the worldwide loss of species and the damage of ecosystems, has continued to accelerate. Studying animal distribution, movement and behavior is of critical importance to address environmental challenges such as spreading of diseases, invasive species, climate and land-use change. Camera traps have proven to be an appropriate technique for continuous animal monitoring in an automated 24/7/52 documentation. However, camera traps including distance measurement are not widely deployed, despite providing valuable additional cues to detect animals (cf. fig. \ref{fig:foreground_background_separation} (top)), to distinguish individual animals in animal hordes (cf. fig. \ref{fig:foreground_background_separation} (bottom)), to locate animals in the observed environment and facilitate the automation of laborious ecological studies like estimating population densities \citep{camera_trap_distance_sampling}.

In computer vision, distance measurements are represented by an additional so-called depth channel in images or video clips. Given Grey-value images as visual output of camera traps, e.g. using infrared cameras for wildlife monitoring at night or twilight (cf. fig. \ref{fig:foreground_background_separation} (top-left)), the additional depth channel captures the distance (cf. fig. \ref{fig:foreground_background_separation} (top-right)). The depth channel is commonly visualized in terms of heatmaps where distance is highest in blue and lowest in red. Color images with an additional depth channel are referenced as RGB-D images where the color components of the images are encoded by three channels of the primary colors red, green and blue while the fourth channel shows the depth information (cf. fig. \ref{fig:foreground_background_separation} (bottom)).

One of the most popular approaches to derive depth information, is stereo vision. Given two cameras, displaced horizontally from each other, two slightly differing perspectives of an observed scene are used to derive the depth of observed scene objects in a way similar to human stereo vision. 

\subsection{Objective and Contributions}
This study proposes an automated camera trap-based approach to detect and identify animals using depth estimation. To detect and identify individual animals, we propose a novel method for the so-called instance segmentation which is deep learning-based technique to detect and delineating each distinct object of interest appearing in an image or a video clip. This novel approach was evaluated in terms of a proof-of-concept in a zoo scenario using an RGB-D camera trap. 

\subsection{Related Work}
\label{sec:RelWork}

To the best of our knowledge, this is the first contribution to apply RGB-D camera traps and RGB-D instance segmentation to wildlife and animal monitoring (cf. surveys of \citet{CameraTrapStudy2020KaysEtAl}, \citet{CameraTrapStudy2020CaravaggiEtAl}, \citet{CameraTrapStudy2019WearnEtAl}, \citet{CameraTrapStudy2018SchneiderEtAl}, \citet{CameraTrapStudyScotsonEtAl2017}, \citet{CameraTrapStudyShengEtAl2014}, \citet{CameraTrapStudyRoveroEtAl2013}, \citet{CameraTrapStudy2008RowcliffeEtAl}).

Previous approaches concerned with instance segmentation of RGB-D images tackle on the one hand more controllable indoor scenarios \citep{CrossModalCVPR, DepthLayering}. Other approaches are highly specialized, e.g. by requiring multiple perspectives on a scene \citep{rgbd_instance_segmentation_2} or being highly reliant on the spatial set-up of the scene observation \citep{rgbd_instance_segmentation_1}.
For this study, the most relevant approach to instance segmentation of RGB-D images is proposed by \cite{ResidualRegret}. This approach has been evaluated on two outdoor datasets and will be discussed some more detail in section \ref{sec:D-MaskRCNN}.

Given conventional color images, i.e. RGB images, Mask R-CNN \citep{maskrcnn} is the current state-of the art for instance segmentation. Its deep learning-based architecture is comprised of a feature pyramid network (FPN) backbone to extract features appropriate for the visual detection of object instances, a region proposal network (RPN) which predicts locations and bounding boxes (BB) of likely instances, a classifier deriving a class probability distribution for each BB region, and a mask header computing an instance mask for each BB region, i.e., a precise shape description of the detected object instances. 

\section{Methods and Material}

\subsection{RGB-D Instance Segmentation}
\label{sec:D-MaskRCNN}
Based on the Mask R-CNN architecture \citep{maskrcnn}, we propose an advanced architecture for applying instance segmentation to RGB-D images that we call \textit{Depth Mask R-CNN} or for short \textit{D-Mask R-CNN} that utilizes additional depth information to improve the prediction of bounding boxes and segmentation masks for the detection and localization of object instances as well as their identification \citep{LabRep}. 

\begin{figure}[H]
    \centering
    \begin{subfigure}[b]{\textwidth}
        \centering
        \includegraphics[width=\textwidth]{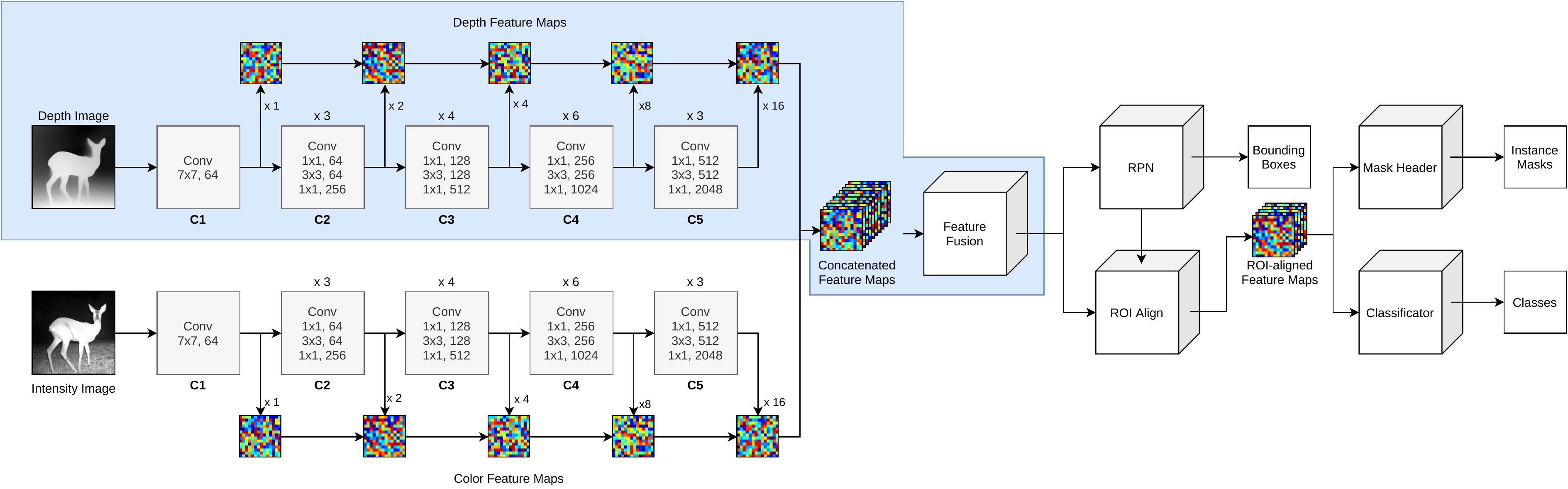}
    \end{subfigure}
    \hfill
    \caption{The architecture of \textit{D-Mask R-CNN}. The blue shaded area depicts contributions to the standard Mask R-CNN architecture. These contributions are explained in detail in this section.}
    \label{fig:mrccn_depth}
\end{figure}

It is an important design decision to build \textit{D-Mask R-CNN} on the basis of the well-developed and popular Mask R-CNN architecture \citep{maskrcnn}. This choice allows (1) to use systems parameters, i.e. model weights, pretrained on huge existing datasets such as ImageNet \cite{imagenet}, (2) to compare performance in presence and absence of depth information and (3) potentially implement existing and proven extensions of Mask R-CNN, such as keypoint detection, in the future. 
The system architecture of \textit{D-Mask R-CNN} is depicted in fig. \ref{fig:mrccn_depth}. 

\paragraph{The depth backbone.} The complete architecture is build on top of the Mask R-CNN implementation in the detectron2 framework \citep{detectron2}. In addition to the existing color image backbone, which is a ResNet-50 model \citep{resnet} pretrained on ImageNet \citep{imagenet}, we employ an almost identical backbone for processing the depth channel.
\vspace*{-1.5ex}
\paragraph{Initialization of depth backbone.} The depth backbone is initialized with the same parameters, i.e., network weights, as the color backbone, except in the first layer. In this layer the weights expect a three-channel RGB color image while the depth channel is just one-dimensional. Therefore, we average over the first weight dimension of the image backbone to obtain the appropriate initial weights in the first layer of the depth backbone. While training, the weights of the depth backbone will of course diverge from the weights of the color backbone, i.e. we employ no weight sharing between both backbones.
Alternatively, we could have also initialized the weights of the depth backbone randomly. But initialization from a pretrained model is desirable because certain filters, e.g. filter specialized for edge detection, are equally desirable and useful for interpreting the depth channel of RGB-D images.
\vspace*{-1.5ex}
\paragraph{Fusion of color and depth features.} When propagating inputs through both backbones, we extract intermediate feature maps of depth 256 at different scales as in the case of a single backbone in the standard Mask R-CNN. 
We then concatenate the feature maps of both backbones at each level (depth 512) and pass them through a single convolutional layer with kernel size $3 \times 3$ (one specialized for each layer) thereby reducing the depth of the concatenated feature maps from 512 back to 256.  We call this operation \textit{feature fusion}, as it fuses the feature information of all three color channels and the depth channel for further processing in coherent way. 
While our D-Mask R-CNN architecture is similar to the approach by \cite{ResidualRegret} in that both employ two separate backbones for the color and depth channels, we impose no prior constraints on the network architecture regarding the selection of features obtained from both backbones.
\vspace*{-1.5ex}
\paragraph{Processing of consolidated color and depth features.} From then on the workflow of the standard Mask R-CNN architecture can be reused - but now processing the consolidated color and depth information. The fused feature maps are given to the region proposal network (RPN) which computes likely instance bounding boxes. The feature maps are then aligned to each bounding box by the region of interest (ROI) alignment. These aligned feature maps are then given to the mask header and classifier to compute instance masks and class predictions, respectively. 


\subsection{Data Material}

Since camera traps including distance sensors are not widely deployed, we evaluate our approach to RGB-D instance segmentation, \textit{D-Mask R-CNN}, with a synthetic dataset comprising RGB-D video clips generated by rendering of synthetic wildlife scenarios. 

To provide a proof-of-concept application, we installed an RGB-D camera trap in a zoo and evaluated \textit{D-Mask R-CNN} on the captured RGB-D video clips. 

\subsubsection{The synthetic Data}

Each animal has an associated running animation which is applied and temporally randomized to sample from all possible states of motion.  We also randomize angle, height and field of view of both the camera and illumination while keeping both roughly pointed in the same direction and in the same place. We then render greyscale, depth, class and instance images using the Blender package \citep{blender}. We render greyscale images instead of color images to simulate camera trap results from nighttime or dawn where infrared sensors generate greyscale images. 

\begin{figure}[H]
    \centering
    \begin{subfigure}[b]{\textwidth}
        \centering
        \includegraphics[width=0.8\textwidth]{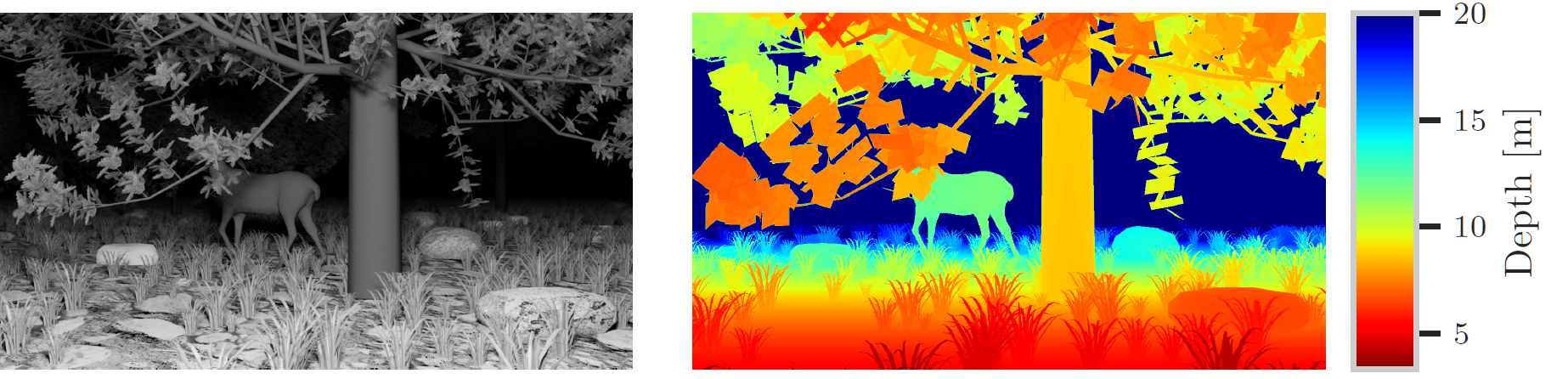}
    \end{subfigure}
    \hfill
    \begin{subfigure}[b]{\textwidth}
        \centering
        \includegraphics[width=0.8\textwidth]{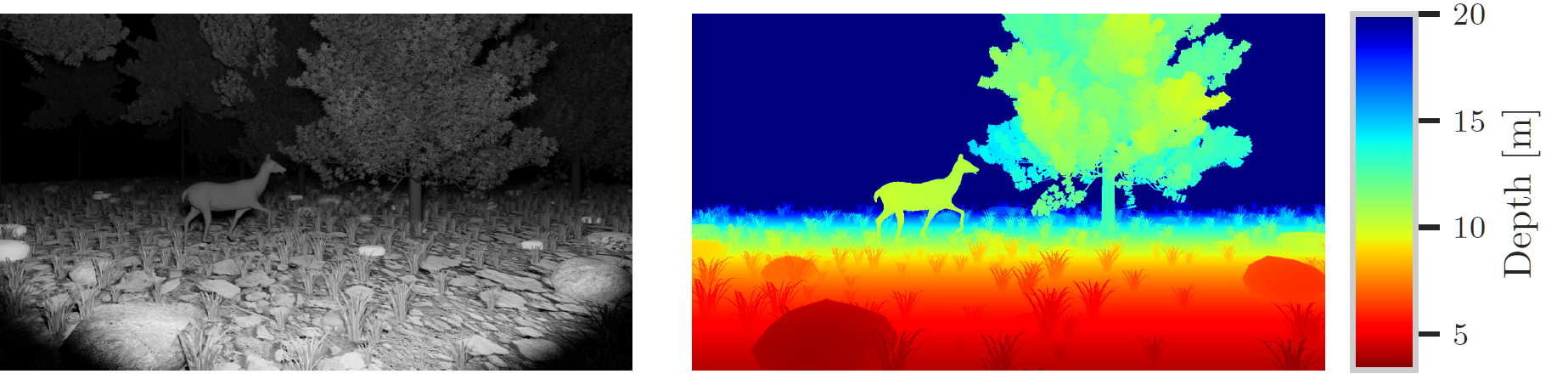}
    \end{subfigure}
    \caption{Two frames from video clips of the synthetic dataset. Left: intensities, right: depth.}
    \label{fig:synthdata}
\end{figure}

We generated RGB-D video clips depicting four animal classes: deer, boar, hare and fox. 
Fig. \ref{fig:synthdata} depicts two frames from video clips of the synthetic dataset.
Table \ref{tab:SynthData} gives an overview of the synthetic dataset.

\begin{table}[H]
\resizebox{\textwidth}{!}{
\begin{tabular}{@{}lcccccccc@{}}
\toprule
 & Split      & \# Frames & \# Deers & \# Boars & \# Hares & \# Foxes & \# Instances & Source of Depth                   \\ \midrule
\multirow{2}{*}{Synthetic}                                                                & train      & 1909      & 2536     & 1212     & 934      & 410      & 5092         & \multirow{2}{*}{Rendered}         \\
  & test & 525       & 630      & 285      & 209      & 121      & 1245         &                                   \\ \bottomrule
\end{tabular}
}
\caption{Statistics of the synthetic dataset.}
\label{tab:SynthData}
\end{table}

\subsubsection{The camera trap dataset} 

\begin{figure}[H]
    \centering
    \begin{annotatedFigure}
        {\includegraphics[width=0.55\textwidth]{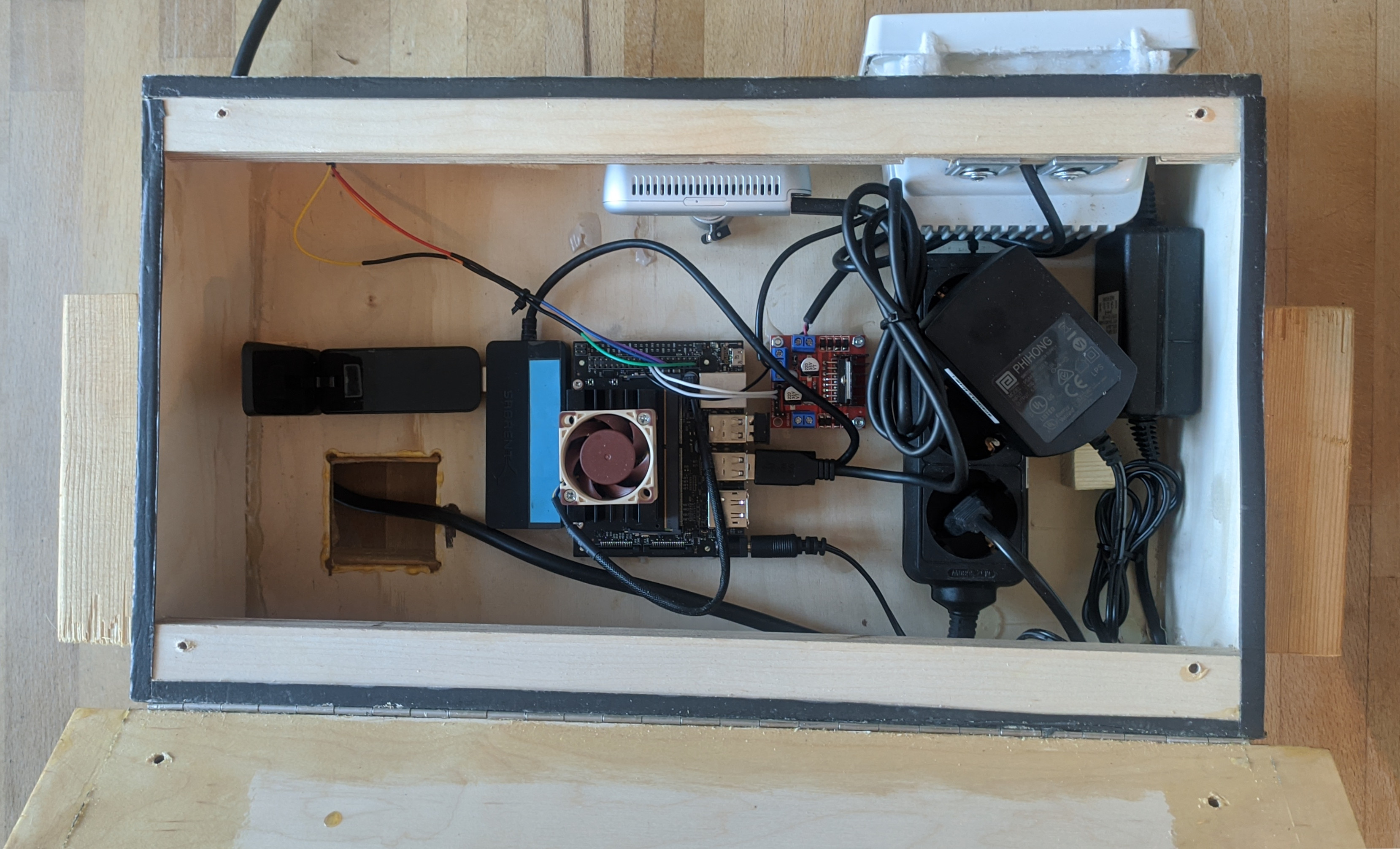}}
        \annotatedFigureBox{0.426,0.7308}{0.586,0.8208}{A}{0.426,0.7308}
        \annotatedFigureBox{0.395,0.3324}{0.542,0.6115}{B}{0.395,0.3324}
        \annotatedFigureBox{0.173,0.672}{0.32,0.9093}{C}{0.173,0.672}
        \annotatedFigureBox{0.545,0.4906}{0.624,0.6114}{D}{0.624,0.4906}
        \annotatedFigureBox{0.599,0.7033}{0.87,0.9907}{E}{0.87,0.7033}
        \annotatedFigureBox{0.162,0.4923}{0.355,0.611}{F}{0.162,0.4923}
    \end{annotatedFigure}
    \caption{Build-up of the RGB-D camera trap. (A): \rs, (B): \jetson, (C) Passive Infrared Sensor (PIR, not directly visible in this image), (D) L298N for controlling (E): Infrared Lamp for nighttime illumination, (F): tp-link Archer T4U Wireless LAN adapter}
    \label{fig:camera_trap_v2_photos_internals}
\end{figure}
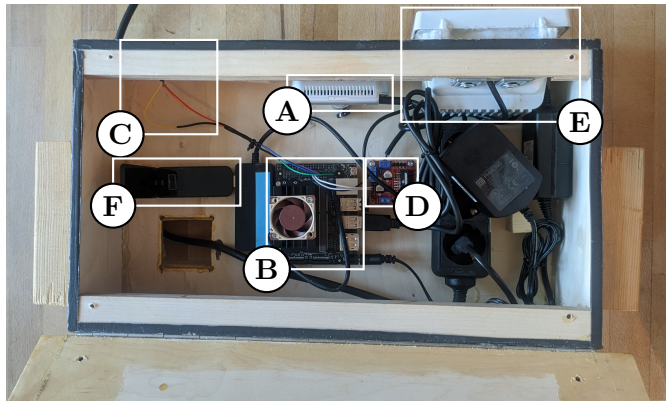

\paragraph{RGB-D camera trap.} We designed and build an RGB-D camera trap by employing low-cost, off-the-shelf components with a special emphasis on the versatility with respect to varying lighting conditions. To accomplish this versatility, we employ an \rs. As an active infrared stereo camera (i.e. two cameras paired with an additional illumination source), it facilitates a wider range of lighting conditions than pure structured light cameras which often fail to find correspondences in brightly-lit scenes. 
The build-up of the RGB-D camera trap is depicted in fig. \ref{fig:camera_trap_v2_photos_internals}. The build-up and its components are explained in detail in this section.

\paragraph{Daytime and nighttime modes.} We switch between daytime and nighttime modes at sunrise and sunset, respectively. Sunrise and sunset times are computed on-device based on the configured latitude and longitude of the camera trap. 

In daytime mode, the camera trap records a color stream from the RGB camera and a depth stream, which is computed on the \rs via active stereo-vision. In strongly illuminated areas, stereo-correspondences are computed passively using the two IR cameras. In weakly illuminated areas, the dot pattern projected by the IR projector are used for additional correspondences. The projected IR pattern is not visible in the color image, as the RGB camera is equipped with an IR filter. 

In nighttime mode, active IR illumination is provided by the IR lamp to achieve sufficient exposure of the IR cameras. 
This illumination 
does not disturb the species of interest. The left IR camera (arbitrary choice) records a greyscale video as substitution for the color video (which is strongly underexposed in absence of visible light). The IR projector is further turned off as its pattern would be visible in the IR video. The correspondences for depth computation are therefore acquired passively. Table \ref{tab:camera_trap_v2_modes} summarizes the differences between the two modes. 

\begin{table}[H]
\centering
\begin{tabular}{@{}lcc@{}}
\toprule
                  & Daytime Mode                        & Nighttime Mode          \\ \midrule
Motion Detection  & Image-based                           & PIR                    \\
Depth Acquisition & Active Stereo                         & Passive Stereo         \\
Image Acquisition & Color Camera                          & Infrared Camera        \\
Illumination      & Ambient Light + Active Stereo Pattern & Infrared Lamp          \\ \bottomrule
\end{tabular}
\caption{Comparison of daytime mode and nighttime mode.}
\label{tab:camera_trap_v2_modes}
\end{table}

\begin{figure}[H]
    \centering
    \begin{tikzpicture}[spy using outlines={rectangle, magnification=4,connect spies}]
        \node [inner sep=0pt] { \includegraphics[width=0.5\textwidth]{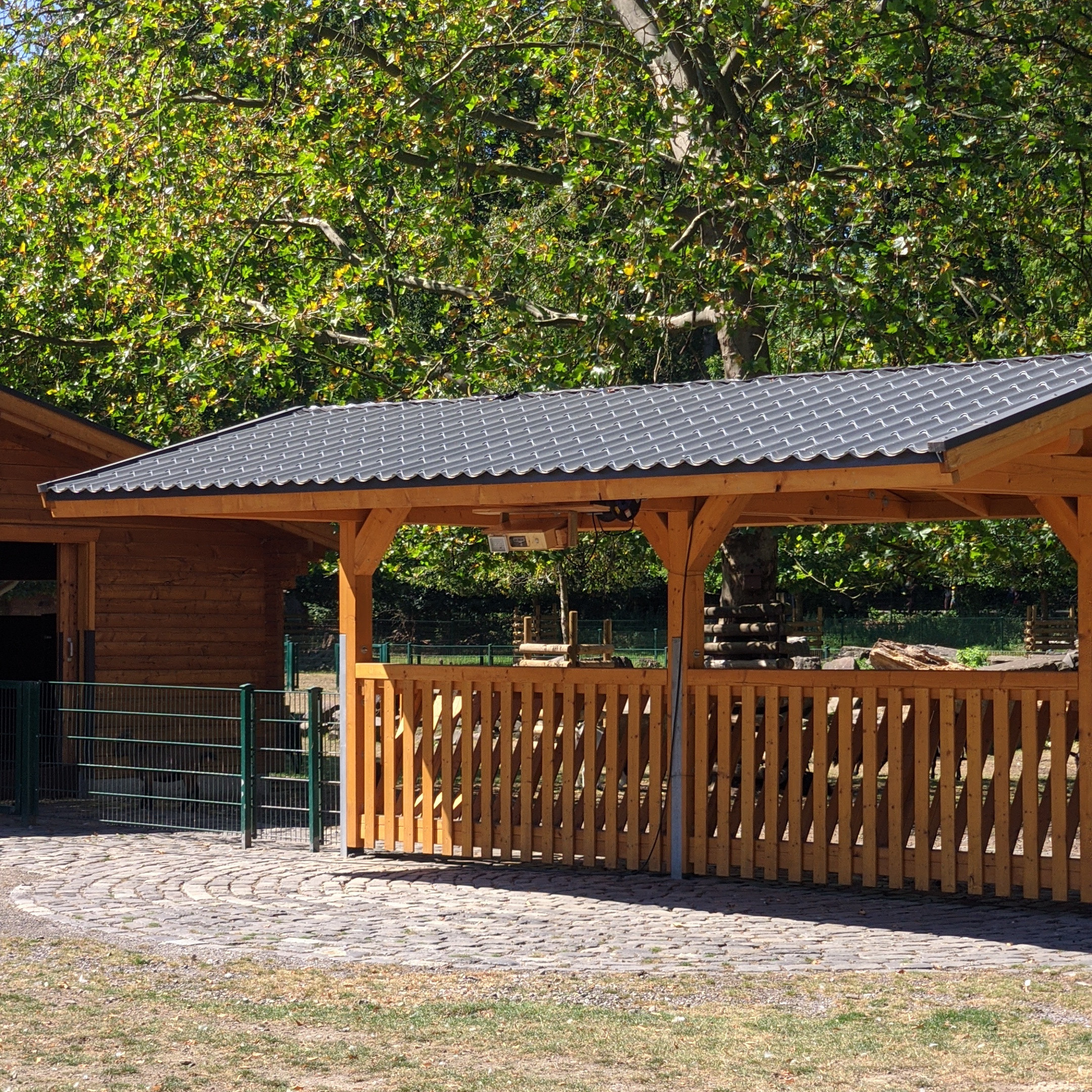} };
        \coordinate (spypoint) at (-0.1,0.1);
        \coordinate (spyviewer) at (2.2,2.2);
        \spy[every spy on node/.append style={thick},white,width=3cm,height=3cm] on (spypoint) in node [fill=white] at (spyviewer);
    \end{tikzpicture}
    \caption{The RGB-D camera trap in the \tierpark.}
    \label{fig:camera_trap_v2_photos_onsite}
\end{figure}

%
\vspace*{-1ex}
\paragraph{Location} We placed the RGB-D camera trap inside an enclosure 
at the \tierpark in Cologne, as depicted in figure \ref{fig:camera_trap_v2_photos_onsite}. It is suspended under the roof of a wooden shelter which provides additional weather protection and places the electronics out of reach of animals. 

\paragraph{Motion Detection}
During daytime, the color camera is used for motion detection using difference images together with a mean color change threshold and background subtraction using Gaussian mixture models \citep{gmm2} together with a foreground ratio threshold. We combine both methods with a preprocessing step which masks out visitor areas. 

During nighttime the active infrared illumination is needed to use the infrared camera motion detection. However, the active infrared illumination would consume large amounts of energy. We therefore employ an additional, digital passive infrared sensor (PIR) to act as motion detector. 

\begin{figure}[H]
    \centering
    \begin{subfigure}[b]{\textwidth}
        \centering
        \includegraphics[width=\textwidth]{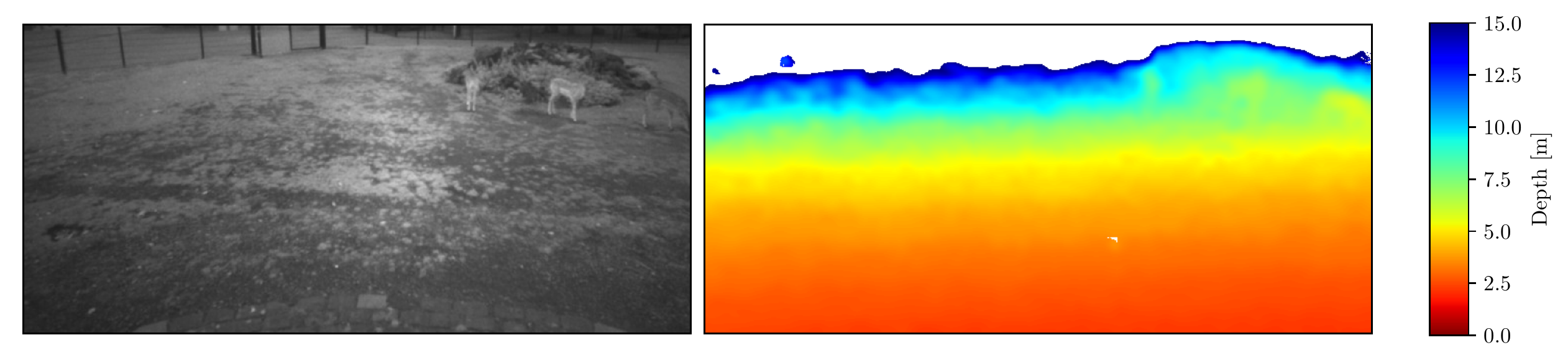}
    \end{subfigure}
    \hfill
    \begin{subfigure}[b]{\textwidth}
        \centering
        \includegraphics[width=\textwidth]{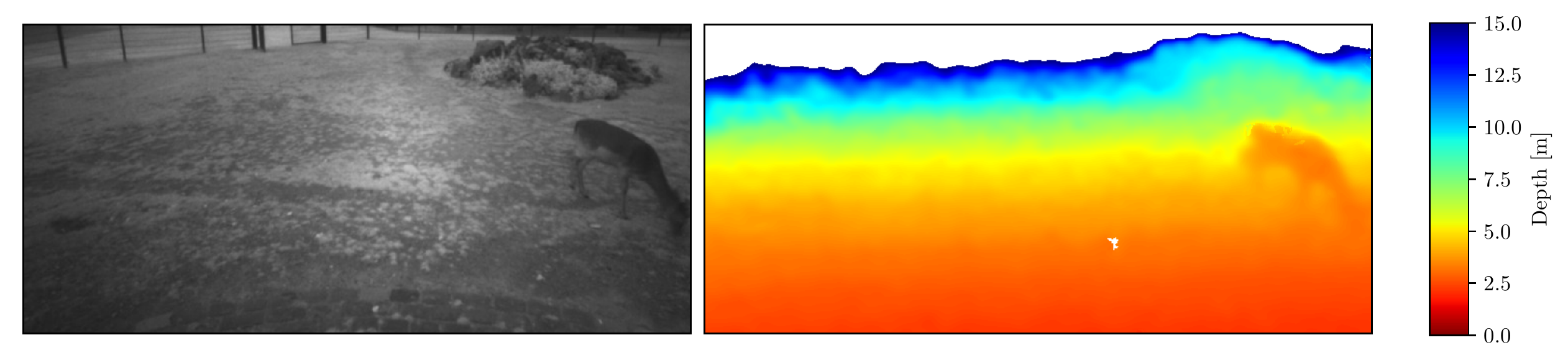}
    \end{subfigure}
    \caption{Two frames from video clips of the RGB-D camera trap dataset. Left: intensities, right: depth. White areas represent missing depth information.}
    \label{fig:rgbddata}
\end{figure}

Fig. \ref{fig:rgbddata} depicts two frames from video clips of the RGB-D camera trap dataset.
Table \ref{tab:TrapData} gives an overview of the RGB-D camera trap dataset.

\begin{table}[H]
\begin{center}
\resizebox{0.8\textwidth}{!}{
\begin{tabular}{@{}lccccc@{}}
\toprule
 & Split      & \# Frames & \# Deers & \# Instances & Source of Depth                   \\ \midrule
\multirow{2}{*}{\begin{tabular}[c]{@{}l@{}}Zoo Lindenthal\\at Night\end{tabular}} & train      & 350       & 462      & 462          & \multirow{2}{*}{Stereo Camera}    \\
& test & 62        & 62       & 62           &                                   \\ \bottomrule
\end{tabular}
}
\end{center}
\caption{Statistics of the camera trap dataset.}
\label{tab:TrapData}
\end{table}

\section{Evaluation}
\label{sec:eval}

\subsection{Instance segmentation with depth and without depth}
\label{sec:SegEval}

Since camera traps including distance sensors are not widely deployed, we evaluate our approach to RGB-D instance segmentation, \textit{D-Mask R-CNN}, with the synthetic dataset comprising RGB-D video clips generated by rendering of synthetic wildlife scenarios. 

As depicted in \ref{tab:SynthData}, 1909 frames of the synthetic dataset have been used for training of \textit{D-Mask R-CNN} while 525 frames have been used, i.e., roughly a train-test split of 0.71:0.29. 

We employ a subset of the official COCO evaluation metrics \cite{coco} which are useful in our target domain: the mean average precision (AP) as an average over 10 IoU (intersection over union) levels, the AP with an IoU threshold of 50\% ($AP_{50\%}$), AP with an IoU threshold of 75\% ($AP_{75\%}$) and AP scores for the observed four different animal classes. The resulting scores of our proposed \textit{D-Mask R-CNN} are compared with the classic Mask R-CNN approach of \citet{maskrcnn}, i.e., Mask R-CNN with and without using depth information in table \ref{tab:instance_segmentation_metrics_synthetic}. Our proposed \textit{D-Mask R-CNN}  clearly outperforms the classic Mask R-CNN in all metrics. 

\begin{table}[H]
\centering
\begin{tabular}{@{}lccc@{}}
\toprule
\multicolumn{1}{l|}{}                    & \multicolumn{1}{c|}{Class}                         & \multicolumn{1}{c|}{\textit{D-Mask R-CNN}} & \multicolumn{1}{c}{Mask R-CNN}          \\ \hline
Bounding Boxes                           & \multicolumn{1}{l}{}                               &                               &                                        \\ \hline
\multicolumn{1}{l|}{AP}                  & \multicolumn{1}{c|}{\multirow{3}{*}{marginalized}} & \multicolumn{1}{c|}{\textbf{47.85\%}}  & \multicolumn{1}{c}{38.04\%}   \\
\multicolumn{1}{l|}{AP50}                & \multicolumn{1}{c|}{}                              & \multicolumn{1}{c|}{\textbf{67.94\%}}  & \multicolumn{1}{c}{51.97\%}   \\
\multicolumn{1}{l|}{AP75}                & \multicolumn{1}{c|}{}                              & \multicolumn{1}{c|}{\textbf{54.99\%}}  & \multicolumn{1}{c}{45.99\%}   \\
\multicolumn{1}{l|}{\multirow{4}{*}{AP}} & \multicolumn{1}{c|}{deer}                          & \multicolumn{1}{c|}{\textbf{63.92\%}}  & \multicolumn{1}{c}{51.50\%}   \\
\multicolumn{1}{l|}{}                    & \multicolumn{1}{c|}{boar}                          & \multicolumn{1}{c|}{\textbf{55.73\%}}  & \multicolumn{1}{c}{42.05\%}   \\
\multicolumn{1}{l|}{}                    & \multicolumn{1}{c|}{hare}                          & \multicolumn{1}{c|}{\textbf{15.46\%}}  & \multicolumn{1}{c}{06.63\%}   \\
\multicolumn{1}{l|}{}                    & \multicolumn{1}{c|}{fox}                           & \multicolumn{1}{c|}{\textbf{56.30\%}}  & \multicolumn{1}{c}{51.97\%}   \\ \hline
Segmentation Masks                             & \multicolumn{1}{l}{}                               &                               &                                        \\ \hline
\multicolumn{1}{l|}{AP}                  & \multicolumn{1}{c|}{\multirow{3}{*}{marginalized}} & \multicolumn{1}{c|}{\textbf{35.49\%}}  & \multicolumn{1}{c}{26.47\%}   \\
\multicolumn{1}{l|}{AP50}                & \multicolumn{1}{c|}{}                              & \multicolumn{1}{c|}{\textbf{63.83\%}}  & \multicolumn{1}{c}{50.28\%}   \\
\multicolumn{1}{l|}{AP75}                & \multicolumn{1}{c|}{}                              & \multicolumn{1}{c|}{\textbf{40.69\%}}  & \multicolumn{1}{c}{25.28\%}   \\
\multicolumn{1}{l|}{\multirow{4}{*}{AP}} & \multicolumn{1}{c|}{deer}                          & \multicolumn{1}{c|}{\textbf{42.78\%}}  & \multicolumn{1}{c}{31.17\%}   \\
\multicolumn{1}{l|}{}                    & \multicolumn{1}{c|}{boar}                          & \multicolumn{1}{c|}{\textbf{50.58\%}}  & \multicolumn{1}{c}{37.18\%}   \\
\multicolumn{1}{l|}{}                    & \multicolumn{1}{c|}{hare}                          & \multicolumn{1}{c|}{\textbf{06.86\%}}  & \multicolumn{1}{c}{03.17\%}   \\
\multicolumn{1}{l|}{}                    & \multicolumn{1}{c|}{fox}                           & \multicolumn{1}{c|}{\textbf{41.75\%}}  & \multicolumn{1}{c}{34.35\%}   \\ \bottomrule
\end{tabular}
\caption{AP scores of the animal detection of \textit{D-Mask R-CNN} on the synthetic dataset for bounding box predictions and segmentation mask predictions. }
\label{tab:instance_segmentation_metrics_synthetic}
\end{table}

\subsection{RGB-D Instance segmentation applied to the camera trap dataset}

\begin{figure}[H]
    \centering
    \begin{subfigure}[b]{\textwidth}
        \centering
        \includegraphics[width=\textwidth]{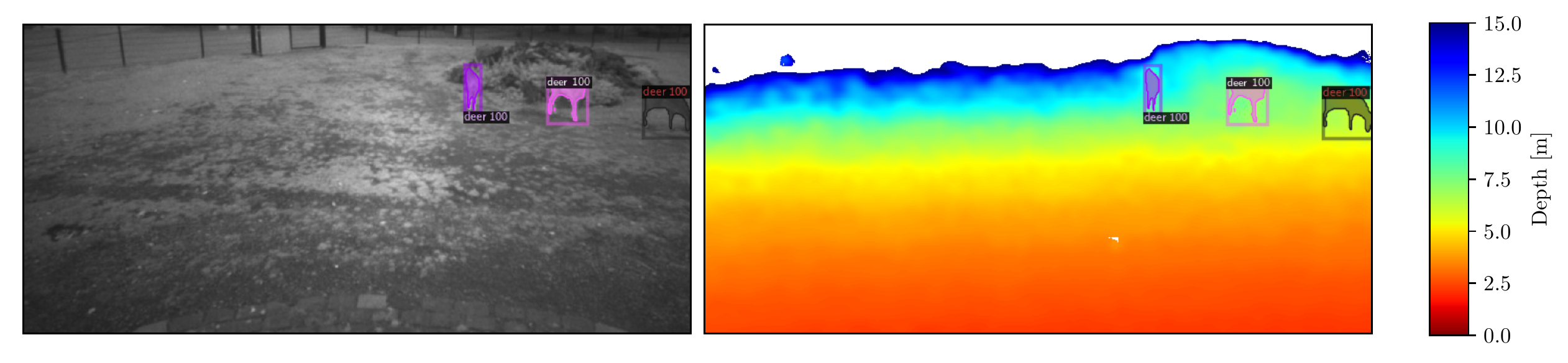}
    \end{subfigure}
    \hfill
    \begin{subfigure}[b]{\textwidth}
        \centering
        \includegraphics[width=\textwidth]{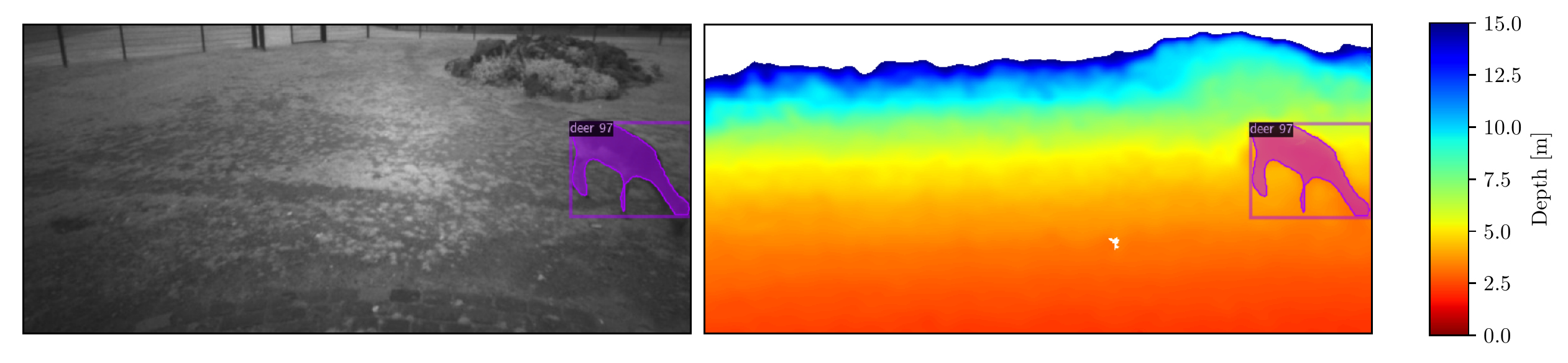}
    \end{subfigure}
    \caption{Two frames from video clips of the RGB-D camera trap dataset overlaid with bounding box predictions and segmentation mask predictions from \textit{D-Mask R-CNN}. Left: intensities, right: depth.}
    \label{fig:PredictionExamples}
\end{figure}

To provide a proof-of-concept application, we applied and evaluated \textit{D-Mask R-CNN} on the captured RGB-D video clips of the RGB-D camera trap installed in the Lindenthal Zoo where the evaluation only take into account the observed deer.  
Figure \ref{fig:PredictionExamples} shows two exemplary results.

\begin{table}[H]
\centering
\begin{tabular}{@{}lcc@{}}
\midrule
Bounding Boxes                           & \textit{D-Mask R-CNN}                             \\ 
\midrule
\multicolumn{1}{l|}{AP}                 & \multicolumn{1}{c}{59.94\%} \\
\multicolumn{1}{l|}{AP50}                & \multicolumn{1}{c}{94.50\%} \\
\multicolumn{1}{l|}{AP75}                 & \multicolumn{1}{c}{63.96\%}
 \\ \midrule
Segmentation                              & \textit{D-Mask R-CNN}   \\ 
\midrule
\multicolumn{1}{l|}{AP}                  & \multicolumn{1}{c}{37.27\%} \\
\multicolumn{1}{l|}{AP50}                & \multicolumn{1}{c}{94.50\%} \\
\multicolumn{1}{l|}{AP75}                & \multicolumn{1}{c}{13.25\%} 
 \\ \bottomrule
\end{tabular}
\caption{AP scores of the animal detection of \textit{D-Mask R-CNN} on the camera trap dataset.}
\label{tab:instance_segmentation_metrics_lindenthal}
\end{table}


\section{Conclusion}

In this study, we propose an automated camera trap-based approach to detect and identify animals using depth estimation. To detect and identify individual animals, we propose \textit{D-Mask R-CNN} for the detection and delineation of distinct animals in RGB-D video clips. An experimental evaluation shows the benefit of the additional depth estimation in terms of improved average precision scores of the animal detection compared to the standard approach that relies just on the image information. \textit{D-Mask R-CNN} shows on a synthetic dataset of animal frames improved AP scores of 47.85\% and 35.49\% for animal detection by bounding boxes and segmentation masks, respectively, compared to the standard Mask R-CNN approach with corresponding AP scores of 38.04\% and 26.47\%, respectively. 

This novel approach \textit{D-Mask R-CNN} was also evaluated in terms of a proof-of-concept in a zoo scenario using an RGB-D camera trap. \textit{D-Mask R-CNN} shows AP scores of 59.94\% and 37.27\% for deer detection by bounding boxes and segmentation masks, respectively.

Future work will expand the RGB-D camera trap dataset for training and testing. Additionally, the deployment of stereo-based RGB-D camera traps with larger baselines is planned to improve depth estimation for more distant animals. 

Last but not least, \textit{D-Mask R-CNN} will be used for automating ecological models such as camera trap distance sampling \citep{camera_trap_distance_sampling} or presence-absence-models \citep{presence_absence}. The modular architecture of \textit{D-Mask R-CNN} also allows to include additional tasks such as keypoint detection that can be used within behavioral studies.

\section*{Acknowledgments}

We gratefully acknowledge the German Federal Ministry of Education and Research (Bundesministerium für Bildung und Forschung (BMBF), Bonn, Gemany (AMMOD - Automated Multisensor Stations for Monitoring of BioDiversity: FKZ 01LC1903B) for funding.

We thank Thomas Ensch, Michael Gehlen and the entire team of the Lindenthaler Tierpark for their cooperation by hosting our experimental camera trap hardware on-site. We thank Alejandro Berni Garcia for his help with the construction of the wooden camera trap casing.


\bibliography{references}

\end{document}